\documentclass{article}

\usepackage[final]{corl_2019}

\usepackage[utf8]{inputenc} 
\usepackage[T1]{fontenc}    
\usepackage{url}            
\usepackage{booktabs}       
\usepackage{amsfonts}       
\usepackage{nicefrac}       
\usepackage{microtype}      

\usepackage{color,xcolor}
\usepackage{epsfig}
\usepackage{graphicx}

\usepackage{adjustbox}
\usepackage{array}
\usepackage{booktabs}
\usepackage{colortbl}
\usepackage{float,wrapfig}
\usepackage{hhline}
\usepackage{multirow}
\usepackage{subcaption} 

\usepackage{amsmath,amsfonts,amsthm,amssymb}
\usepackage{bm}
\usepackage{nicefrac}
\usepackage{microtype}

\usepackage{changepage}
\usepackage{extramarks}
\usepackage{fancyhdr}
\usepackage{lastpage}
\usepackage{setspace}
\usepackage{soul}
\usepackage{xspace}

\usepackage{url}

\usepackage{algorithm, algorithmic}
\usepackage{enumerate}
\usepackage{todonotes} 

\usepackage{titlesec}
\usepackage{makecell}

\usepackage{amsmath,amsthm,amssymb}
\usepackage{mathtools}

\newcolumntype{L}[1]{>{\raggedright\let\newline\\\arraybackslash\hspace{0pt}}m{#1}}
\newcolumntype{C}[1]{>{\centering\let\newline\\\arraybackslash\hspace{0pt}}m{#1}}
\newcolumntype{R}[1]{>{\raggedleft\let\newline\\\arraybackslash\hspace{0pt}}m{#1}}

\newcommand{\sect}[1]{Section~\ref{#1}}

\newcommand{\eqn}[1]{Equation~\ref{#1}}
\newcommand{\fig}[1]{Figure~\ref{#1}}
\newcommand{\tbl}[1]{Table~\ref{#1}}

\newcommand{\ignore}[1]{}

\makeatletter
\DeclareRobustCommand\onedot{\futurelet\@let@token\@onedot}
\def\@onedot{\ifx\@let@token.\else.\null\fi\xspace}

\makeatother

\definecolor{MyDarkBlue}{rgb}{0,0.08,1}
\definecolor{MyDarkGreen}{rgb}{0.02,0.6,0.02}
\definecolor{MyDarkRed}{rgb}{0.8,0.02,0.02}
\definecolor{MyDarkOrange}{rgb}{0.40,0.2,0.02}
\definecolor{MyPurple}{RGB}{111,0,255}
\definecolor{MyRed}{rgb}{1.0,0.0,0.0}
\definecolor{MyGold}{rgb}{0.75,0.6,0.12}
\definecolor{MyDarkgray}{rgb}{0.66, 0.66, 0.66}

\newcommand{\myitem}{\vspace{-4pt}\item}
\titlespacing*{\section}{0pt}{0pt plus 2pt minus 2pt}{0pt plus 2pt minus 2pt}
\titlespacing\subsection{0pt}{0pt plus 1pt minus 1pt}{0pt plus 1pt minus 1pt}
\DeclarePairedDelimiterX{\infdivx}[2]{(}{)}{%
  #1\;\delimsize|\delimsize|\;#2%
}

\def\x{\mathbf{x}}
\def\t{\tau}
\def\st{\tilde{\tau}}

\title{Model-Based Planning with Energy-Based Models}

\author{
  Yilun Du\\
  MIT CSAIL\\
  \texttt{yilundu@mit.edu} \\
  \And
  Toru Lin\\
  MIT CSAIL\\
  \texttt{torulk@mit.edu} \\
  \And
  Igor Mordatch \\
  Google Brain \\
  \texttt{imordatch@google.com}
}

\begin{document}

\maketitle

\begin{abstract}

Model-based planning holds great promise for improving both sample efficiency and generalization in reinforcement learning (RL). We show that energy-based models (EBMs) are a promising class of models to use for model-based planning. EBMs naturally support inference of intermediate states given start and goal state distributions. We provide an online algorithm to train EBMs while interacting with the environment, and show that EBMs allow for significantly better online learning than corresponding feed-forward networks. We further show that EBMs support maximum entropy state inference and are able to generate diverse state space plans. We show that inference purely in state space - without planning actions - allows for better generalization to previously unseen obstacles in the environment and prevents the planner from exploiting the dynamics model by applying uncharacteristic action sequences. Finally, we show that online EBM training naturally leads to intentionally planned state exploration which performs significantly better than random exploration.
\end{abstract}

\section{Introduction}

Recent advances in reinforcement learning have primarily relied on model-free approaches, and have shown strong performance across a wide range of domains ~\citep{Silver2016Mastering, Mnih2013Playing, kempka2016vizdoom,dota,andrychowicz2018learning}. However, several issues persist in model-free approaches: they have poor sample efficiency~\citep{Sutton1990Integrated}, and are unable to adapt to new tasks or domains~\citep{nichol2018gotta}. A key reason for these problems is that in model-free approaches, an agent's policy and its knowledge of environmental dynamics are entangled.

Model-based approaches, on the other hand, separate learning of dynamics model from learning of policies. This means that the dynamics model learned in one domain can then be extracted and applied to other domains and tasks \citep{du2019task}, suggesting the better generalizability of model-based methods. We therefore adopt such an approach in this paper, combining planning with learned dynamics models to accomplish a wide variety of tasks.

Most approaches towards planning with dynamics models consider models whose next states are conditioned on both actions and current states. However, such approaches limit models to environments in which the same set of actions are used. Furthermore, planning in action space can lead to the planner exploiting the dynamics models by applying uncharacteristic action sequences that take the agent outside the competence region of the model \citep{levine2018reinforcement}. In contrast, learning a dynamics model that directly predicts next-state dynamics both generalizes to different action spaces and is less likely to explore risky states. However, directly predicting next state dynamics is difficult, as probability distributions over the next states are difficult to model due to their multi-modal nature. We find the energy-based models (EBMs) are naturally able to capture multi-modal distributions and exhibit maximum entropy sampling of next states that construct diverse plans to reach the goals.

For such approaches to work in an real-world robot learning settings, models must be learned in an online fashion, where dynamics can change abruptly and be heavily correlated. As was also observed in \citep{du2019implicit} for generative modeling tasks, we find that EBMs are particularly suited to learning in online settings. They significantly outperform corresponding feed-forward networks benchmarks and exhibit robustness to heavily correlated experience.

Furthermore, for online model learning to perform well, it must be able to effectively explore the surrounding environment. We find that training EBMs naturally leads to good exploration that significantly outperforms random exploration.

Our overall contributions in this work are three-fold. First, we propose a framework for online learning with EBMs for planning and show that they perform well under online model learning. Second, we show EBMs support the control as inference framework and allow maximum entropy reinforcement learning on next states as opposed to actions, allowing diverse state based planning that shows better generalization than planning conditioned on both states and actions. Finally, we show that online planning with EBMs naturally gives rise to exploration.

\begin{figure}
\begin{center}
\begin{subfigure}{\textwidth}
\includegraphics[width=1.0\linewidth]{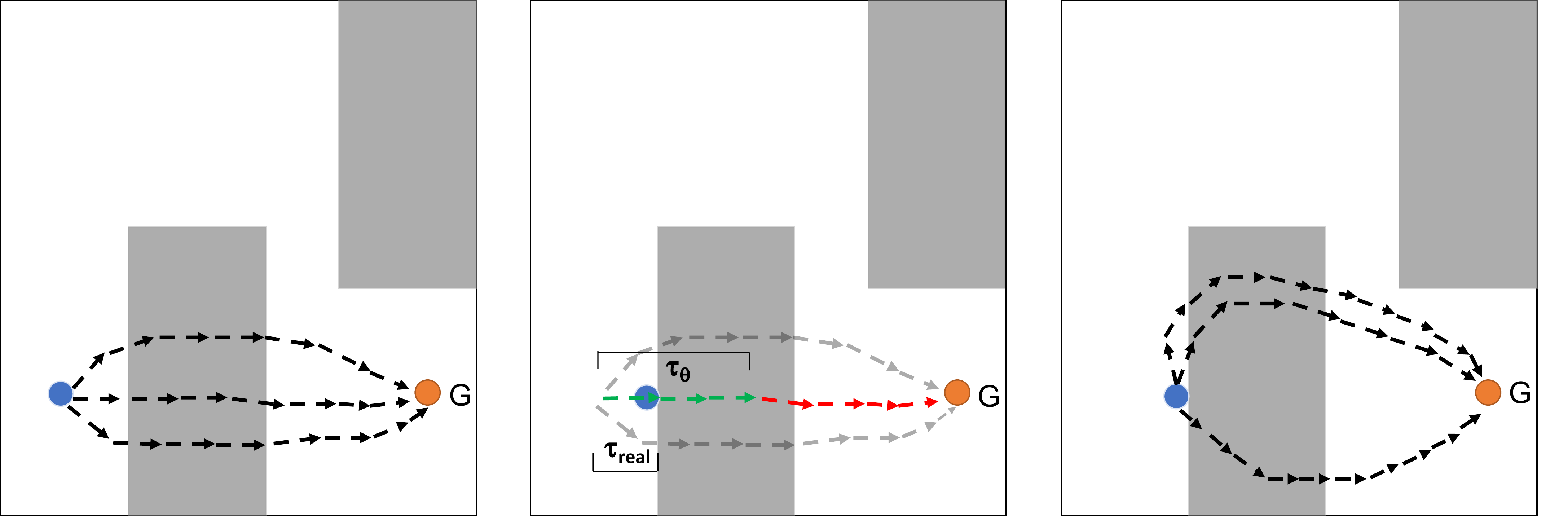}
\end{subfigure} 
\end{center}
\caption{\small Overview of online training procedure with a EBM where grey areas represent inadmissible regions. Plans from the current observation to goal state is inferred from the EBM (left). A particular plan is chosen and executed until a planned state deviates significantly from an actual state (middle). The EBM is then trained on all real transitions $\tau_{\text{real}}$ and all planned transitions before the deviation (in green) $\tau_{\theta}$, while transitions afterwards (red) are ignored. A new plan is then generated from the new location of the agent (right)}
\label{fig:overview}
\vspace{-10pt}
\end{figure}
\section{Related Work}

A number of model classes have been explored for model-based planning in robotics and artificial intelligence literature, from feed-forward, recurrent \citep{nagabandi2018neural}, temporal segment \citep{mishra2017prediction} and Bayesian \citep{gal2016improving} neural networks, to locally-linear models \citep{yip2014model, mordatch2016combining} and Gaussian processes \citep{deisenroth2011pilco}. By contrast, energy-based models have been an under-explored for model-based planning and it is our aim to showcase the favorable properties this model class exhibits. While work of \citep{haarnoja2017reinforcement} explored energy-based models of policies for model-free reinforcement learning, we instead use them to model environment dynamics in a model-based setting.

Energy-based models have seen success in other applications, such as natural image modeling \citep{du2019implicit, dai2019exponential}. These works noted EBMs for favorable compositionality and online learning properties, which we take advantage of in this work. A number of methods have been used for inference and sampling in EBMs (or equivalently, planning), from Gibbs Sampling \citep{hinton2006fast}, to Langevin Dynamics \citep{du2019implicit}, and learned samplers \citep{kim2016deep}. We have not focused on the choice of sampler/planner in this work, and found method of \citep{williams2017model} to work well. Other planning methods, such as those based on direct collocation methods \citep{mordatch2012discovery, erez2012trajectory} can potentially be used instead.

A common approach to achieve exploration behavior in reinforcement learning has been to use explicit rewards, known as intrinsic motivation \citep{oudeyer2009intrinsic, schmidhuber2010formal}. Examples include rewarding empowerment, information gain about model of the dynamics \citep{Pathak2017Curiosity}, or state space coverage \citep{houthooft2016vime, tang2017exploration}. Maximum entropy models are another approach to induce exploratory behavior \citep{haarnoja2018soft} and what we rely on this our work as well. We show that contrastive training of EBMs is particularly conducive to exploration.

\section{Model-based Planning with Energy-Based Models}

In this section, we describe the overall approach towards model-based planning with EBMs. We reformulate planning as inference over a graphical model defined by a composition of EBMs. Inference over the graphical model can be seen as maximum entropy reinforcement learning (see \citep{levine2018reinforcement} for a review). We first give a background overview of relevant terminology and EBMs, then introduce the graphical model formulation and finally discuss online training of EBMs.

\subsection{Energy-Based Models and Terminology}

Consider a standard Markov Decision Process (MDP) represented by the tuple $\langle S, A, T, R\rangle$, where $S$ is the set of all possible state configurations, $A$ is the set of actions available to the agent, $T$ is the transition distribution, and $R$ is the reward function. Under this setup, define a state transition pair $(s_{t}, s_{t+1})$ between the states at timesteps $t$ and $t+1$. Define $E_\theta(s_{t}, s_{t+1}) \in \mathbb{R}$ as the energy function, which we parameterize with a deep neural network.  We interpret the energy function as unnormalized probability distribution over state transition by defining the distribution as $p_{\theta}(s_{t+1}| s_{t}) \propto e^{-E_\theta(s_{t}, s_{t+1})}$.

To sample from the defined probability distribution, we use Model Predictive Path Integral (MPPI) algorithm \citep{williams2017model}, which is shown to converge to the full posterior distribution in \citep{williams2017information}. The mathematical formulation is shown below, where $x := (s_{t}, s_{t+1})$:
\begin{align}
\quad \Tilde{x}^k = \sum_i w_i x_i^k, \quad x_i^k \sim N(x^{k-1}, \sigma), \quad w_i = \left(\frac{e^{-E_{\theta}(x_i^k)}}{\sum_j e^{-E_{\theta}(x_i^j)}}
\right)
\label{eq:mppi}
\end{align}
Other valid inference algorithms that sample from the posterior are also applicable to EBMs, such as Langevin Dynamics \citep{du2019implicit} or Hamiltonian Monte Carlo (HMC).

We note that this form of sampling makes it easy to add additional constraints to a probability distribution by simply adding the constraint as an energy. To train an EBM, we follow the methodology defined in \citep{du2019implicit}. We train models by minimizing 
\begin{align}
\mathbb{E}_{x^+ \sim p_D}{ E_\theta(x^+)} - \mathbb{E}_{x^- \sim p_\theta}{ E_\theta(x^-)}.
\end{align}
Intuitively, doing so decreases the energies of observed transitions (i.e. more likely transitions), and decreases the energies of transitions sampled from the model's distribution (i.e. less likely transitions).

\subsection{Planning with Energy-Based Models}

In the previous subsection, we described a way to learn state transition models $p_{\theta}(s_{t+1}|s_t)$. We now discuss how to use models to do inference over trajectories. Given a learned model $p_{\theta}(s_{t+1}|s_{t})$, we can model the likelihood of a trajectory $\tau$ under the model as a product of factors
\begin{align}
  p_{\theta}(\tau) &= p_{\theta}(s_1, s_2, \ldots, s_T) = \prod_{t=1}^{T-1} p_{\theta}(s_{t+1}|s_{t})  \\
 & \propto \exp (-\sum_{t=1}^{T} E(s_{t}, s_{t+1}))
\end{align}

We can likewise do inference across this product using MPPI. Note that we directly sample states rather than actions in our formulation. We generate temporally smooth trajectory perturbations following approach of \citep{kalakrishnan2011stomp} (where the last two rows of the finite difference matrix A are removed to allow end states of trajectories to be unconstrained).

Given a particular fixed goal state $g$ and start state $s_1$, we can do inference over intermediate states by sampling from the probability distribution among state transitions between $s_2$, ..., $s_T$
\begin{align}
p_{\theta}(s_2, \ldots, s_T| s_1, g) \propto \exp (-\sum_{t=1}^{T-1} E(s_{t}, s_{t+1}) - E(s_T, g)) \label{eqn:infer}
\end{align}
to get a plan. Alternatively, instead of using a fixed goal state $g$, we can represent the goal state with a Gaussian distribution around it, $P(g)$, and similarly perform inference over 
\begin{align}
p_{\theta}(s_2, \ldots, s_T| s_1, g) \propto \exp (-\sum_{t=1}^{T-1} E(s_{t}, s_{t+1}) - (s_T - g)^2) \label{eqn:gauss}
\end{align}
to get a plan. We found that inference over both distributions worked well using MPPI. Throughout our experiment, we follow \eqn{eqn:gauss} to specify a Gaussian distribution around all goal states used in our experiments, to accommodate additional constraints more flexibly and refer to the resulting state distribution as $p_{\theta}(\tau | s_1, G)$. Actions are not inferred in this sampling process, but are fed into a ground truth inverse dynamics model that given a pair of candidate states outputs an action, but show in \sect{sect:online} that using a learned model also works well.

We can also represent the goal state as the trajectory that has the highest conditional probability of reaching an optimal reward, where the event of reaching an optimal reward is defined as $O_t$ and the probability $P(O_t|s_t)$ is defined as $e^{R(s_t)}$ for a reward function $R(s)$. Inference can then be done on
\[p_{\theta}(\tau| O_{1:T}) \propto \exp (R(s_1)-\sum_{t=1}^{T-1} (E(s_{t}, s_{t+1}) - R(s_{t+1}))) \]
which \citet{levine2018reinforcement} interpret as maximum entropy reinforcement learning on the model. However, while the form proposed in \citep{levine2018reinforcement} considers maximum entropy over actions given a state, we consider maximum entropy of the next state given the current state.

\subsection{Online Learning with Energy-Based Models}

The previous sections described inference done by EBMs pre-trained with generated data. We now turn to the question of how to generate the training data in an on-going manner to simultaneously learn the EBM as the robot operates in the environment. We discuss online training methods of EBMs -- i.e. how to effectively obtain data on state transitions and learn an energy function given a MDP environment represented as a tuple $\langle S, A, T, R\rangle$ and a Goal $G$.

In this setup, we first generate a $T$-step trajectory $\tau_{\theta}$ from the model $p_{\theta}(\tau, s_1, G)$ and use an inverse dynamics model to compute the corresponding actions $a_t$ at each time step. We then execute each action $a_i$ in the real environment to generate another $T$-step trajectory $\tau_{\text{real}}$, stopping prematurely if the real observations deviate significantly from model predictions. After that, we train an EBM to increase the energy of each attempted transition in $\tau_{\theta}$ (imagined transitions) while decreasing energy of real transition in $\tau_{\text{real}}$ (real transitions). We note that perfect planning will have no effect on the model training since $\tau_{\theta} = \tau_{\text{real}}$. For stability, we maintain a replay buffer of past experiences and simulated trajectories. \fig{fig:overview} provides an overview of the process.

Intuitively, our training procedure allows our model to learn a good likelihood distribution over states that the model has observed. However, since we terminate model's planning after significant deviation between real observations and the enacted plan, the model is not trained to minimize the probability of transitions among faraway states. These transitions are thus free to vary over time throughout training, which eventually provides incentive for the model to explore the whole state space. In our experiments section, we illustrate this effect and show that EBMs lead to good exploration. 

For completeness, we include pseudo-code for online training of EBMs, where $\Omega(\cdot)$ denotes a collation operator that converts a trajectory to pairs of state transitions.

\begin{algorithm}
\begin{algorithmic}
    \STATE \textbf{Input:} goal state $G$, step size $\lambda$, number of steps $K$, 
    \STATE number of plan steps $T$, inverse dynamics model $ID$, replay buffers $\mathcal{B}_{pos}$, $\mathcal{B}_{neg}$
    \STATE $\mathcal{B}_{pos} \gets \varnothing$
    \STATE $\mathcal{B}_{neg} \gets \varnothing$
    
    \FOR{environment timestep $i$}
     \STATE \emph{$\triangleright$ Initialize trajectory as a smooth trajectory at start state}
    \STATE $\st^0_i = s_0$ 
    
    \STATE \emph{$\triangleright$ Generate sample from $p_{\theta}(\tau | s_i, G)$ via MPPI}
    \FOR{sample step $k = 1$ to $K$}
    \STATE $\st^k = \sum_i w_i \st^k_i, \quad \st^k_i \sim N(\st^{k-1}_i, \Sigma), \quad w_i = \left(\frac{e^{ (\sum_{i=0}^{T-1} E_\theta (s^k_i, s^k_{i+1})) + (s^k_T - G)^2}}{\sum_j e^{ (\sum_{i=0}^{T-1} E_\theta (s^j_i, s^j_{i+1})) + (s^j_T - G)^2} }
\right) $ 
    \ENDFOR
    
    \STATE $ a \gets ID(\st^K)$ 
    \STATE $\t^+_i \sim \text{simulate environment with actions } a$

    \STATE $\x^+ = \Omega(\t^+) \cup \text{sample}( \mathcal{B}_{pos})  $
    \STATE $\x^- = \Omega(\st^k) \cup \text{sample} (\mathcal{B}_{neg}) $
    \STATE \emph{$\triangleright$ Optimize objective $\mathcal{L}_{2} + \mathcal{L}_\text{ML}$ wrt. $\theta$:} \vspace{1mm}
    \STATE $\Delta \theta \gets \nabla_\theta \frac{1}{N} \sum_i  E_\theta(\x^+_i)^2 + E_\theta(\x^-_i)^2 + E_\theta(\x^+_i) - E_\theta(\x^-_i) $
    \STATE Update $\theta$ based on $\Delta \theta$ using Adam optimizer  \vspace{2mm}
    
    \STATE $\mathcal{B}_{pos} \gets \mathcal{B}_{pos} \cup \x^+$
    \STATE $\mathcal{B}_{neg} \gets \mathcal{B}_{neg} \cup \x^-$
    \ENDFOR
  \end{algorithmic}
 \caption{Online training of an EBM}
 \label{alg:complete}
\end{algorithm}

\section{Experiments}

We perform empirical studies to answer the following questions: Firstly, can EBMs be applied to model learning in an online setting? Secondly, can EBMs be successfully used for maximum entropy inference and what advantages does that carry? Lastly, what exploration behavior and properties do EBMs exhibit?

\subsection{Setup}
\begin{figure}
\begin{center}
\includegraphics[width=1.0\linewidth]{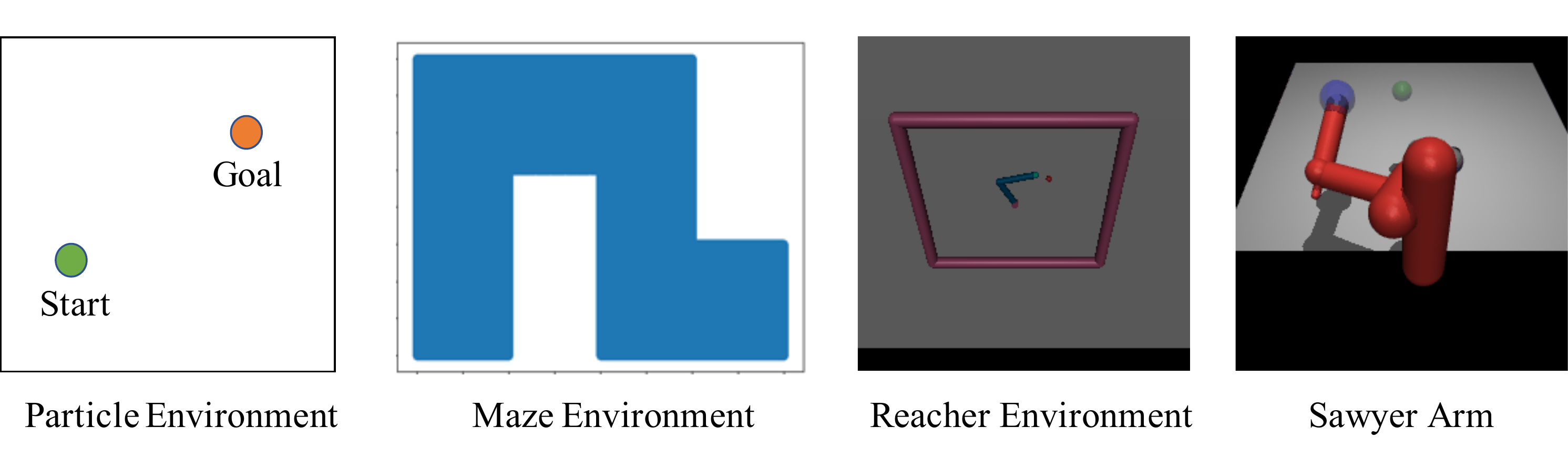}
\end{center}
\caption{\small Illustrations of the 4 evaluated environments. Both particle and maze environments have 2 degree of freedom for x, y movement. The Reacher environment has 2 degrees of freedom corresponding to torques to two motors. The Sawyer Arm environment has 7 degrees of freedoms corresponding to torques.}
\label{fig:enviroment}
\vspace{-10pt}
\end{figure}

We perform experiments on four different environments listed below, with corresponding visualizations in \fig{fig:enviroment}:

\begin{enumerate}
    \myitem \textbf{Particle}: An environment in which a particle is spawned at a start position and must navigate to a goal position; each position is represented by an (x, y) tuple. The observation is the current position of particle, and there are two degrees of freedom that correspond to x-displacement and y-displacement. Reward at each timestep corresponds to negative distance from current position to goal position. Agents are able to move in 0.05 uniform ball around their current location, with size of the map being 2 by 2.
    \myitem \textbf{Maze}: Same setup as the particle environment, but certain areas contain walls that prevent movement of particle.
    \myitem \textbf{Reacher}: The Reacher environment in \citep{Brockman2016OpenAI}. The system consists of two degrees of freedom for angles of joints. The observation is the current rotations and angular velocities of joints. Reward at each timestep corresponds to negative distance from current rotations of joints to target rotations of joints.
    \myitem \textbf{Sawyer Arm}: A simulation of the Sawyer Arm in Mujoco \citep{Todorov2012MuJoCo}. The system is second order and contains 7 degrees of freedom. The observation is the position and velocities of each of the joints, as well as the current end-effector finger position. Reward at each timestep is the negative distance between current end-effector finger and target end-effector finger position. Target end-effector position is either fixed or randomized.
\end{enumerate}

 For each task, we compare our model's performance with a learned deterministic feedforward network (Action FF) that predicts the next state from the current state/action (with the same architecture as an EBM). We generate plans by sampling over states using MPPI, with score calculated from the L2 distance between final and goal state. On the Sawyer Arm task, we further compare our performance with a model-free baseline PPO \citep{Schulman2017Proximal}, using the implementation provided in \citep{baselines}.

We investigate difference in performance of models that have been trained using two different methods, where sources and availability of data are varied. In the case where data is available in advance, models are trained on 100,000 action-state transitions pre-generated from random sampling in each environment. In the case where only online data is available, models are trained by samples generated from interacting with an environment from start state; the training algorithm is outlined by Algorithm \ref{alg:complete}, with replay buffers used on both models.


\subsection{Online Model Learning}
\label{sect:online}
\tbl{tbl:tbl_quant} shows the performance of an EBM compared to action FF on the Particle, Maze and Reacher tasks. First, we compare both methods given a large pre-generated dataset of random interactions; we find that EBM performs slightly better than Action FF. However, when we compare both methods under an online setting, we find that EBM performs \textbf{significantly} better than Action FF. For example, an EBM only experiences a drop of 15.24 in score when switched to the online setting, compared to the score drop of 844.56 experienced by an Action FF model on the online setting. 
\begin{figure}[H]
    \centering
    \begin{minipage}{0.45\textwidth}
        \includegraphics[width=0.8\linewidth]{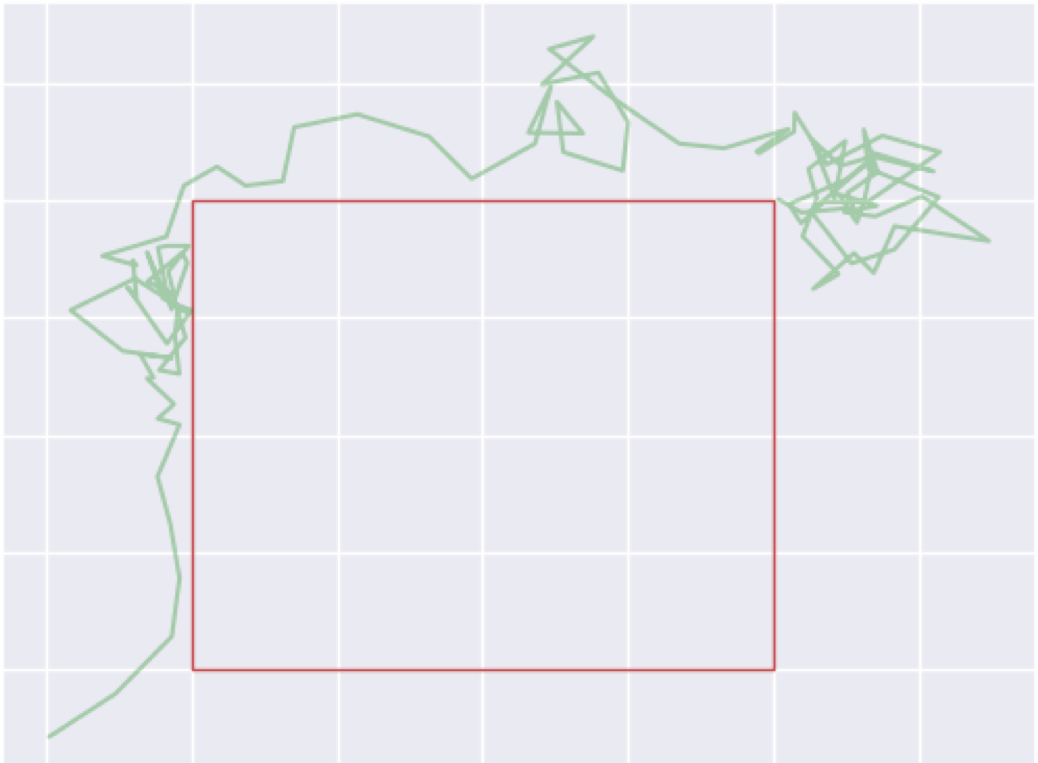}
        \caption{\small Navigation path with a central obstacle the model was not trained with.}
        \label{fig:obstacle}
    \end{minipage}\hfill
    \begin{minipage}{0.45\textwidth}
        \centering
        \begin{subtable}{\textwidth}
        \resizebox{\textwidth}{!}{
        \begin{tabular}{lcccc}
        \toprule
        Data & Model & Particle & Maze & Reacher\\
        \midrule
        \multirow{2}{*}{Pretrained} & EBM & \textbf{-5.14} & -72.07 & \textbf{-19.38} \\
        & Action FF & -6.11 & \textbf{-65.06} & -25.54 \\
        \midrule
        \multirow{2}{*}{Online} & EBM & \textbf{-20.38} & \textbf{-162.97} & \textbf{-29.87} \\
        & Action FF & -850.67 & -949.99 & -42.37 \\
        \bottomrule
        \end{tabular}
        }
        \end{subtable}
        \caption{\small Performance on Particle, Maze and Reacher environments where dynamics models are either pretrained on random transitions or learned via online interaction with the environment. Action FF: Action Feed-Forward Network.}
        \label{tbl:tbl_quant}
    \end{minipage}
    \vspace{-10pt}
\end{figure}

\begin{table}[t]
\centering
\resizebox{\textwidth}{!}{
    \begin{tabular}{lccccc}
    \toprule
     Model & Pretrained (random) & Pretrained (directed) & Pretrained (directed + sequential) & Online (Fixed) & Online (Variable) \\
    \midrule
    EBM & -9569 & -4438 & -5114 & -3782 &  -3907 \\
    Action FF & -10326 & -5041 & -12838 & -9360 & -11942 \\
    \bottomrule
    \end{tabular}
    }
\caption{\small Comparison of performance on the Sawyer Arm environment between Action FF and EBM. In the pretraining setting, we compare models trained using random transitions, directed transitions from an EBM, and directed transitions with correlated data. In the online setting, models are trained on 50,000 simulations of the environment. We find that EBMs perform well online.}
\label{tbl:tbl_sawyer}
\vspace{-10pt}
\end{table}
\begin{figure}
\begin{center}
\begin{subfigure}{\textwidth}
\includegraphics[width=1.0\linewidth]{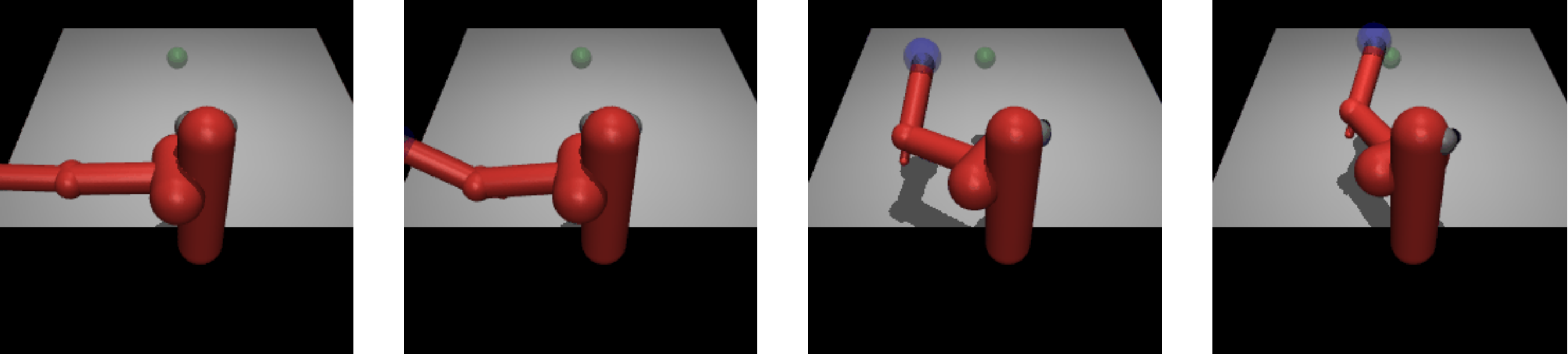}
\end{subfigure} 
\end{center}
\caption{\small Qualitative image showing EBM successfully navigating finger end effector to goal position.}
\label{fig:sawyer_execution}
\vspace{-10pt}
\end{figure}

\tbl{tbl:tbl_sawyer} shows the performance of EBM compared to action FF on the Sawyer arm scenario. We find that in such a setting, using a large pre-generated dataset of random interactions led to insufficient state coverage. To mitigate this, we construct a directed dataset of 100,000 frames from an EBM trained on Sawyer Arm task. With directed data, we find that a pretrained EBM performs slightly better than Action FF, obtaining scores of -4438 and -5041 respectively. In the online training scenario (with either fixed or varied goals), however, we find that EBM performs \textbf{significantly} better (with a score of -3782) than Action FF (with a score of -9360). We show images of execution in \fig{fig:sawyer_execution}. 

On this task, the model free algorithm PPO obtains performance of -9300 with the same amount of experience, and requires 250,000 to 500,000 frames (5 - 10 times more than used in online training of the models) to achieve comparable performance to online training of an EBM. With 50 - 100 times more experience, PPO is able to obtain better scores of -1000 (note that since PPO does not have acceleration priors we do, it is allowed to reach the goal faster thus producing a higher reward - our method moves slower, but both methods successfully reach the goal). Furthermore, PPO is not able to operate in an online manner and does not exhibit zero-shot generalization results of our model, both of which are important in real-world robot learning regimes.

\tbl{tbl:tbl_sawyer} also considers another scenario in which goals are varied across the table. In this setting we find that EBMs still perform better (with a score of -4547 while Action FF obtains a score of (-11942). Furthermore, we can actually apply a model trained on a fixed goal, and generalize to variable goals, and still obtain a score of -4547.

We also ablate dependence on ground truth inverse dynamics. Using recursive least squares \citep{mordatch2016combining} to infer inverse dynamics also leads to good performance of -4694. We find that our learned state distributions is not significantly impacted by inverse dynamics inference, as long as action inference does not suffer from mode collapse (which can occur from neural networks based approaches).

To ablate the effect of exploration and the ability of EBMs to learn models online, in \tbl{tbl:tbl_sawyer} we train both Action FF and EBM models on the directed dataset, but with batches sequentially sampled from the dataset, with each datapoint repeated 100 times without shuffling to mimic the correlated experiences seen during online training. Under this setting, the performance of EBM drops slightly by 676, but performance of Action FF drops catastrophically by 7797 and the model fails to train. When training on static data-set, we do not use a replay buffer of past transitions.

Our results indicate that EBMs are able to learn well online, which is an important and necessary characteristic for models to learn in the real world.

\subsection{Maximum Entropy Inference}
\begin{figure}
\begin{center}
\begin{subfigure}{0.5\textwidth}
\centering
\includegraphics[width=0.7\linewidth]{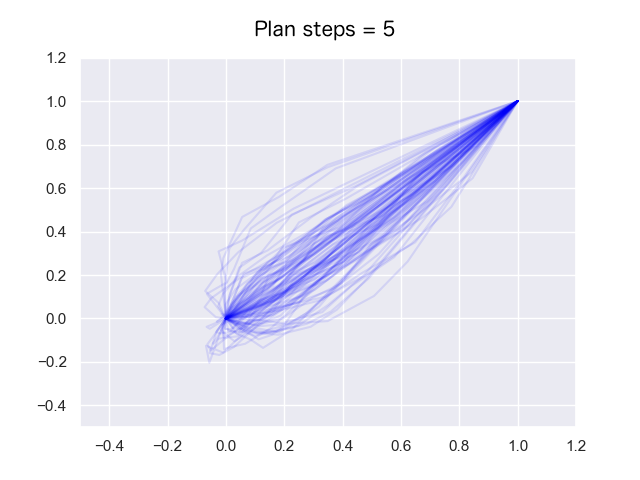}
\end{subfigure}%
\begin{subfigure}{0.5\textwidth}
\centering
\includegraphics[width=0.7\linewidth]{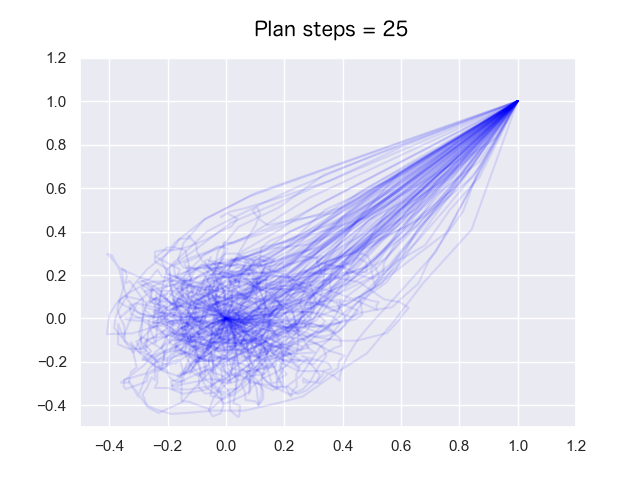}
\end{subfigure}
\end{center}
\caption{\small Effects of varying number of planning steps to reach a goal state. As the number of steps of planning increases, there is a larger envelope of explored states.}
\label{fig:plan_steps}
\vspace{-5pt}
\end{figure}
\begin{figure}
\begin{center}
\includegraphics[width=0.8\linewidth]{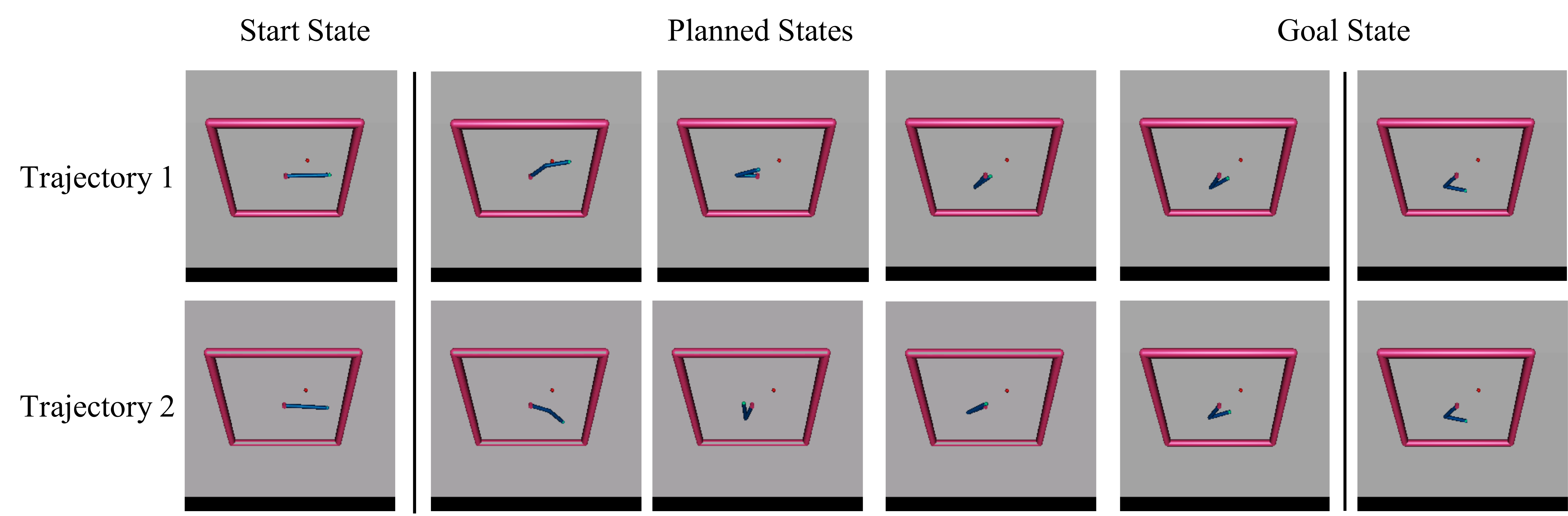}
\end{center}
\caption{\small Illustrations of two different planned trajectories from start to goal in the Reacher environment.}
\label{fig:reacher_plan}
\vspace{-10pt}
\end{figure}

While maximum entropy reinforcement learning has focused on maximizing entropy of actions given a state, sampling from an EBM corresponds to directly maximizing entropy over the next state. In \fig{fig:plan_steps}, we find EBM sampling is capable of generating diverse plans that go from a given start state to goal state. In \fig{fig:plan_steps}, we show that given a fixed start state and end goal, increased number of planned steps leads to a larger envelope of possible trajectories. The same diagram also shows that our method is able to sample across a wide range of trajectories that are different from each other. In \fig{fig:reacher_plan}, we show that in the Reacher environment, we are able to make valid plans with both clockwise and counter-clockwise  given a start and goal state.

We illustrate the power of diverse plans by comparing the generalization performance between planning conditioned on only state space (EBM planning) and planning conditioned on both state and action space (action-conditional planning). In the particle environment, at test time we add a large obstacle not seen during training as the particle attempts to navigate from start state to goal state, as shown in \fig{fig:obstacle}; an EBM is able generalize better obtained a reward of -61.94 while an Action FF obtains a reward of -81.24. 

\begin{figure}
\begin{center}
\includegraphics[width=0.9\linewidth]{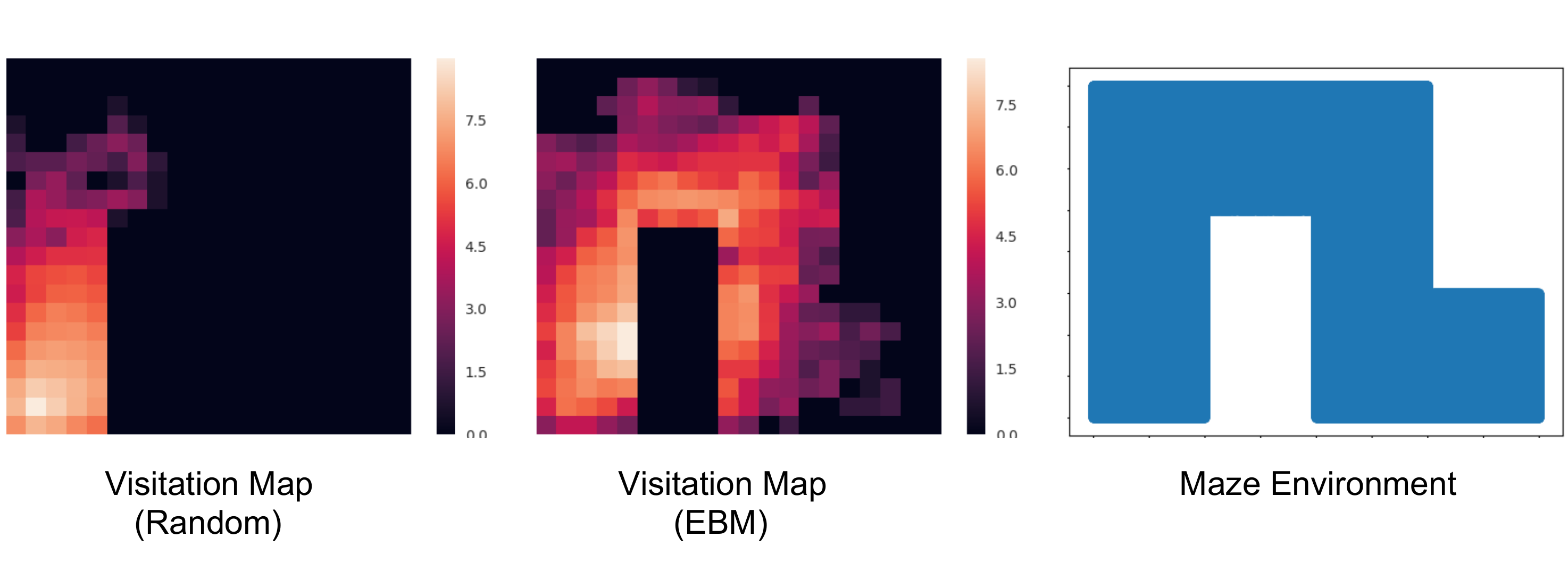}
\end{center}
\caption{\small Illustration of exploration in a maze under random actions (left) as opposed to following an EBM (middle). Areas in blue in the maze environment (right) are admissible, while areas in white are not.}
\label{fig:maze_exploration}
\vspace{-10pt}
\end{figure}
\subsection{Exploration}

\begin{figure}[h]
\begin{center}
\includegraphics[width=0.9\linewidth]{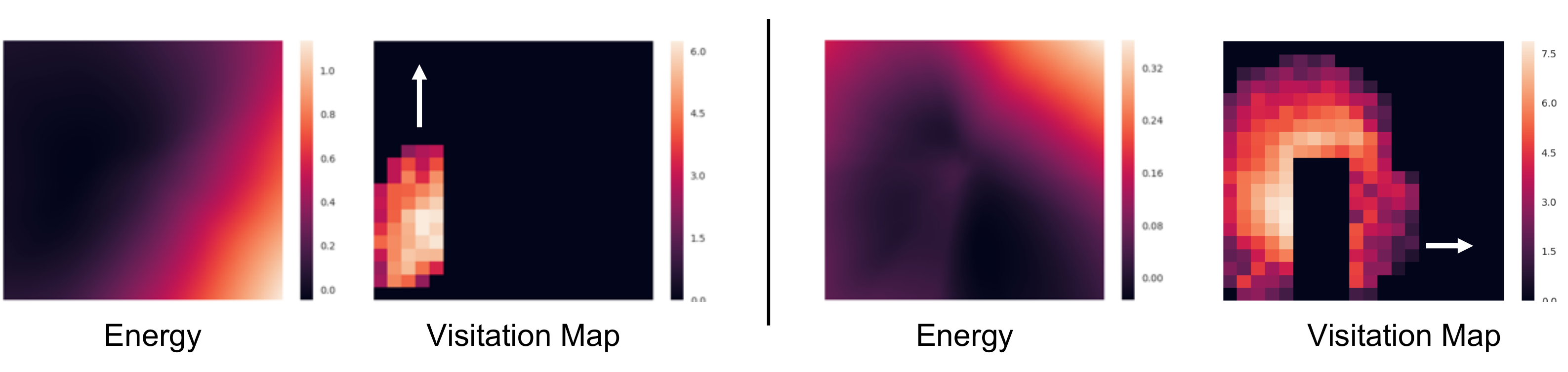}
\end{center}
\caption{\small Illustration of energy values of states (computed by taking the energy of a transition centered at the location) and corresponding visitation maps. While an EBM learns a probabilistic model of transitions in areas already explored, energies of unexplored regions fluctuate throughout training, leading to a natural exploration incentive. Early on in training (left), the EBM puts low energy on the upper corner, incentivizing agent exploration towards the top corner. Later on in training (right), an EBM puts low energy on the right lower corner, incentivizing agent exploration towards the bottom corner.}
\label{fig:energy_visitation}
\vspace{-10pt}
\end{figure}

\begin{wrapfigure}{l}{0.5\textwidth}
\begin{center}
\includegraphics[width=0.8\linewidth]{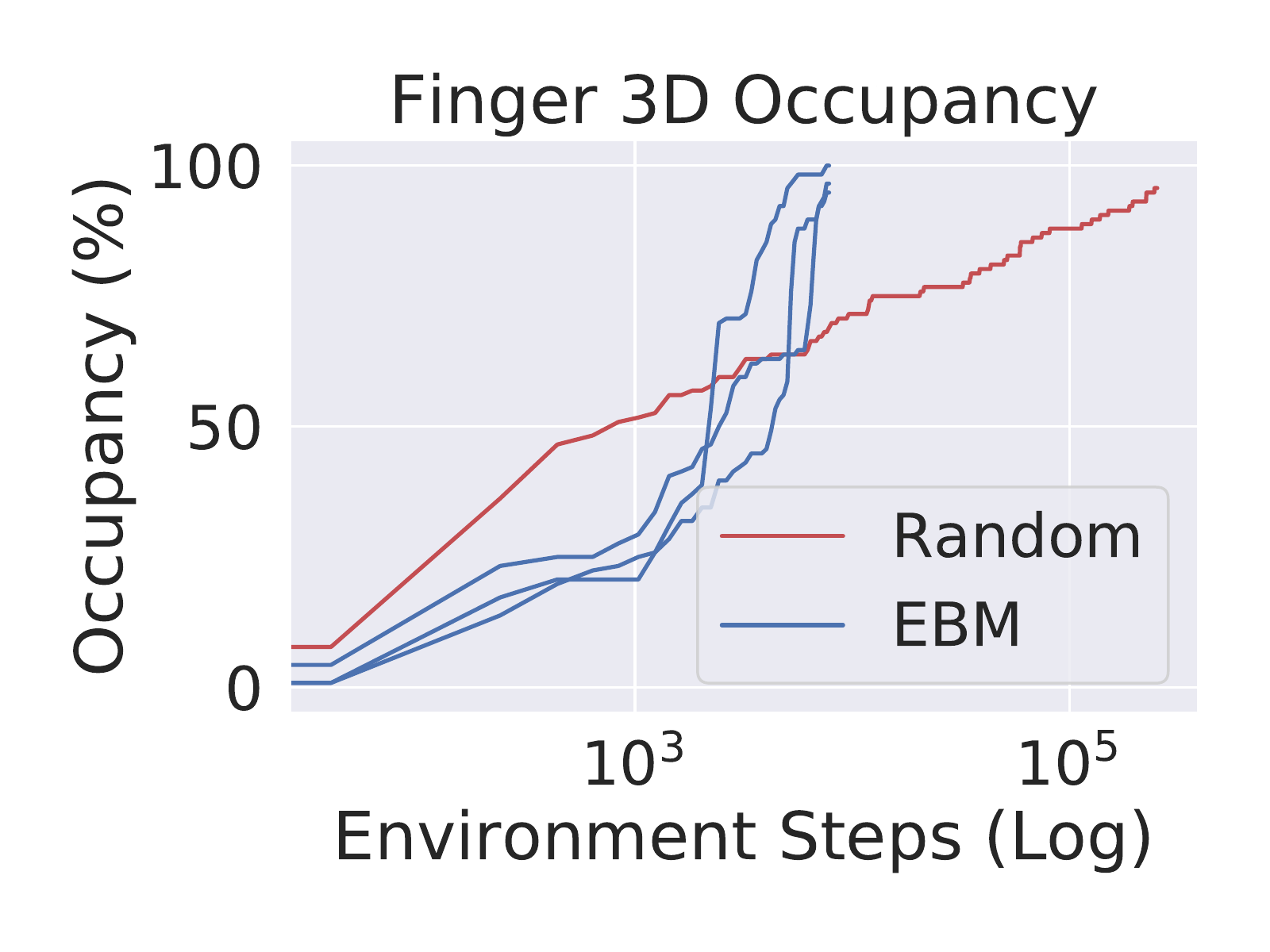}
\end{center}
\caption{\small Comparison of 3D spatial occupation of the finger end-effector of the Sawyer Arm, using random exploration versus using an EBM without a goal on a \textbf{log} scale across 4 different seeds. An EBM allows more directed exploration and explores more states. For the random policy to reach maximum occupancy, more than 200,000 transitions are required.}
\label{fig:finger_exploration}
\vspace{-10pt}
\end{wrapfigure}

We show that an EBMs model naturally incentive exploration. In \fig{fig:maze_exploration} we compare the exploration behavior of an EBM without a goal and a random action agent in the Maze environment. In the time it takes a random policy to explore a hallway of a maze, an EBM is able to explore the entirety of the maze. Similarly, we consider 3D occupancy of the finger end-effector in the Sawyer arm; we define 3D occupancy by partitioning space into 3D voxels and measuring the number of voxels that a finger ends up in. We empirically to be found that the maximum system occupancy was 116. We find in \fig{fig:finger_exploration} that EBMs reaches maximum system occupancy significantly faster compared to random exploration across 4 different seeds. Without a goal, an EBM is able to navigate the arm freely, compared to a random policy that struggles and takes over 100 more times the number of environmental transitions (i.e. over 200,000) to reach maximum occupancy.

We reason that the exploration behavior in EBMs comes from the fact that they learn local dynamics of the world only in the regions that have been explored. This allows the EBMs to assign arbitrary energies to transitions among unexplored states. Values of these energies vary over the course of training, and lead the EBMs to generate plans to reach different unseen states until more of the environment is explored. We illustrate this result in \fig{fig:energy_visitation}, where we show that an EBM puts low energy in a swath of states that are unexplored but reachable in two stages of training, incentivizing exploration of those states while maintaining correct energies for states that have already been explored.

EBMs learn local dynamics models since they are trained on real data transitions and transitions from planning; a plan is followed until it deviates significantly from real transitions. Thus both sets of transitions are consisted of states that are locally close to states an EBM has learned. In contrast, traditional likelihood models for modeling trajectory lower the likelihood of all unseen trajectories, including at unseen states; as a result, planning using such models is unable to explore as adequately.








\section{Discussion}

We have presented some preliminary results on using EBMs for planning. We show that EBMs are a promising class of models for formulating planning as inference. We further show that EBMs behave well under online model learning, and are able to naturally incentivize exploration. We hope to inspire further investigation towards using EBMs for model-based planning.
\newpage
{\small
\bibliography{reference,ebm}
}

\end{document}